\pdfoutput=1

\documentclass[11pt]{article}

\usepackage[]{EACL2023}

\usepackage{times}
\usepackage{latexsym}

\usepackage[T1]{fontenc}

\usepackage[utf8]{inputenc}

\usepackage{microtype}

\usepackage{inconsolata}

\usepackage{graphicx}
\usepackage{url}
\usepackage{latexsym}
\usepackage{makecell}
\usepackage{epstopdf}
\usepackage{tipa}
\usepackage{hyperref}
\usepackage{xstring}
\usepackage{amsfonts}
\usepackage{amsmath}
\usepackage{fixltx2e}
\usepackage[T1]{fontenc}
\usepackage{booktabs}
\usepackage{arabtex}
\usepackage{hhline}
\usepackage{pdfpages}
\usepackage{utf8}
\newcommand{\hide}[1]{}

\newcommand{\AHAMZAUP}{{\^{A}}}

\newcommand{\TAMARBUTA}{{$\hbar$}}

\usepackage{multirow}
\usepackage{subfigure}
\usepackage{amssymb}
\usepackage[super]{nth}
\usepackage[export]{adjustbox}
\usepackage{wasysym}
\usepackage{footmisc}

\definecolor{orange}{HTML}{ffb000}
\definecolor{blue}{HTML}{648fff}
\newcommand{\APGC}{{\sc APGC}}

\title{The User-Aware Arabic Gender Rewriter}

\author{Bashar Alhafni, Ossama Obeid, Nizar Habash \\
  Computational Approaches to Modeling Language Lab\\
  New York University Abu Dhabi\\
  \texttt{\{alhafni,oobeid,nizar.habash\}@nyu.edu}
  }

\setcode{utf8}
\vocalize

\begin{document}
\maketitle
\begin{abstract}
We introduce the User-Aware Arabic Gender Rewriter, a user-centric web-based system for Arabic gender rewriting in contexts involving two users.\footnote{Demo: \url{http://gen-rewrite.camel-lab.com/}} The system takes either Arabic or English sentences as input, and provides users with the ability to specify their desired first and/or second person target genders. The system outputs gender rewritten alternatives of the Arabic input sentences (or their Arabic translations in case of English input) to match the target users' gender preferences.

\end{list}
\end{abstract}


\section{Introduction}
\label{sec:intro}

\begin{figure}[t]
\centering
\includegraphics[width=\linewidth]{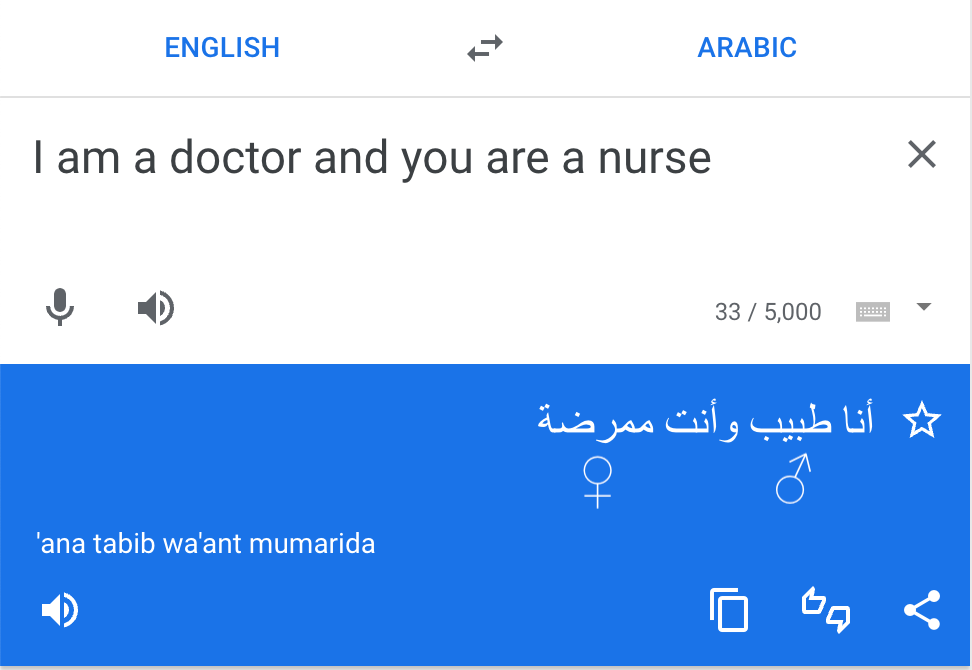}
\caption{\href{https://translate.google.com/}{Google Translate's} output for \textit{``I am a doctor and you are a nurse}'' in Arabic. Doctor is translated to the  masculine form (`\<طبيب>' \textit{Tbyb}), whereas nurse is translated to the feminine form (`\<ممرضة>' \textit{mmrD\TAMARBUTA}).}
\label{fig:google-translate}
\end{figure}

Gender stereotypes, both negative and positive, are manifest in most of the world’s languages \cite{maass1996language,menegatti2017gender} and are further propagated and amplified by NLP systems \cite{sun-etal-2019-mitigating,blodgett-etal-2020-language} (see Figure~\ref{fig:google-translate}).
This is because NLP systems rely on human-created language corpora that mirror the societal biases and inequalities of the world we live in \cite{boyd-crawford-2012,olteanu-2019}. For instance, Figure~\ref{fig:baba-advice}(a) presents part of a cooking recipe published on an Arabic popular cooking website targeting female readers,\footnote{\url{https://www.atyabtabkha.com/}} whereas Figure~\ref{fig:baba-advice}(b) shows part of an article on career advice that is published on Harvard Business Review in Arabic targeting male readers.\footnote{\url{https://hbrarabic.com/}} However, even if overt gender biases are removed from datasets before using them to build NLP models, this will not ultimately reduce the biases produced by systems that are designed to generate a single text output without taking their target users' gender preferences into consideration.

Some commercial NLP systems have solved this problem by generating more than one gender-specific output when the system encounters ambiguous scenarios. For instance, Google Translate generates both feminine and masculine translations when translating gender-neutral English sentences (e.g., \textit{I am a doctor}) to a limited number of languages, such as  Spanish~\cite{Kuczmarski:2018:Approach,Johnson:2020:Approach}. However, this approach does not work well in multi-user contexts (first and second persons, with independent grammatical gender preferences), particularly
when dealing with gender-marking morphologically rich languages. One example of this phenomenon is the Arabic machine translation of the sentence \textit{I am a doctor and you are a nurse}. Figure~\ref{fig:google-translate} shows that Google Translate outputs the Arabic translation \<أنا طبيب وأنت ممرضة> \textit{{\AHAMZAUP}nA Tbyb w{\AHAMZAUP}nt mmrD\TAMARBUTA}\footnote{Arabic HSB transliteration \cite{Habash:2007:arabic-transliteration}.} `I am a [male] doctor and you are a [female] nurse', whereas a more suitable output would include all four possible Arabic translations. 

\begin{table*}[ht!]
    \centering
        \begin{tabular}{cc}
            (a) & (b) \\
            \includegraphics[width=0.44\textwidth]{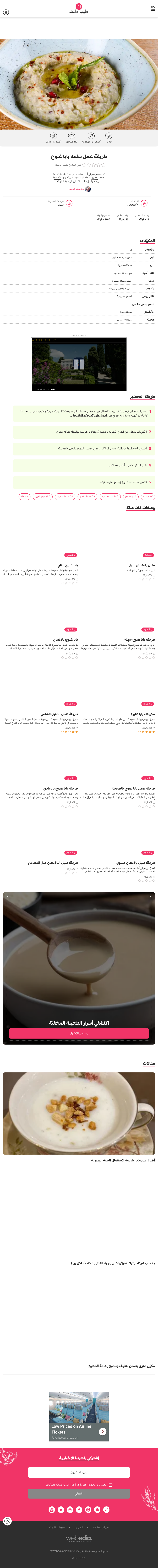} &
            \includegraphics[width=0.44\textwidth]{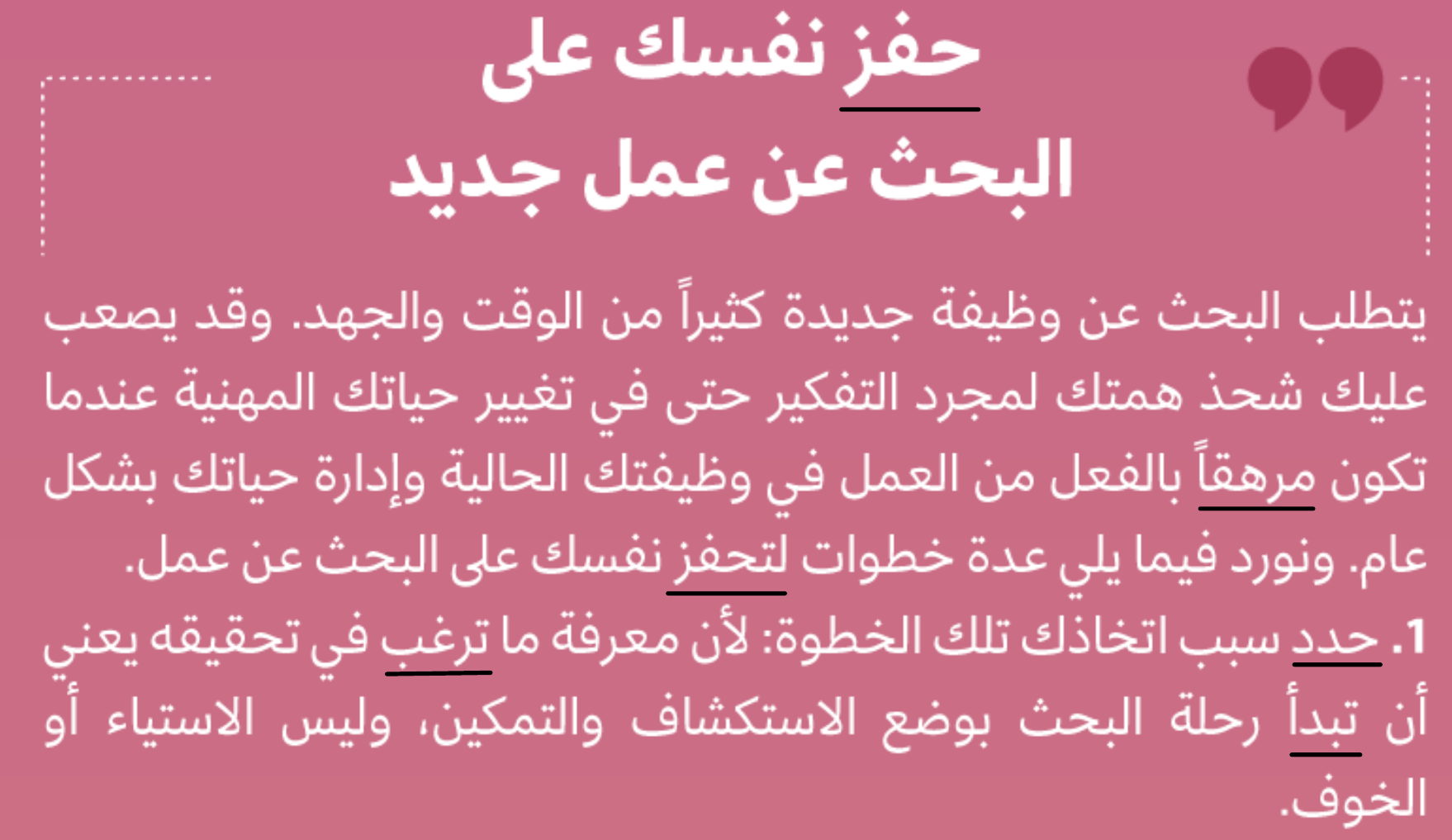}
        \end{tabular}
    \captionof{figure}{Examples of gender-specific text in the wild. Figure (a) is an example of text targeting female readers from a website about cooking recipes. The example is the introduction to a recipe for \href{https://en.wikipedia.org/wiki/Baba_ghanoush}{Baba Ghannouj}. Figure (b) is an example of text targeting male readers from a website about career advice. The example is about an advice on how to find a new job. The underlined words are morphologically marked for the second person feminine in (a), and the second person masculine in (b).
}
    \label{fig:baba-advice}
\end{table*}

One approach to mitigate the ambiguity is to provide the users with the ability to specify their desired target gender preferences so that NLP systems would generate personalized unbiased outputs. To this end, we build on the work of \newcite{alhafni-etal-2022-user} where they formally introduced the task of gender rewriting and developed a user-centric gender rewriting model for Arabic.\footnote{\url{https://github.com/CAMeL-Lab/gender-rewriting/}\label{alhafni22}} We introduce the User-Aware Arabic Gender Rewriter, a user-centric web-based system for Arabic gender rewriting in contexts involving two users. Our system takes either Arabic or English sentences as input, and provides users with the ability to specify their desired first and/or second person grammatical target
genders. The system outputs gender rewritten alternatives of the Arabic input sentences (or their Arabic translations in case of English input) to match the target users’ gender preferences. To the best of our knowledge, this is the first open-access web-based system for Arabic gender rewriting.

Our goal behind creating an easy-to-use web-based multi-user Arabic gender rewriting tool is to enable users to rewrite any Arabic text based on their grammatical gender preferences that are consistent with their social identities. This reduces the gender bias that is caused by user-unaware NLP systems and increases the inclusiveness of Arabic NLP applications, leading to a better user experience.  We envision a future in which websites such as those in Figure~\ref{fig:baba-advice} could use automatic gender rewriting that fits the private preferences of their readers, or that is adjusted  with simple website controls comparable to selecting different languages. 



The rest of this paper is organized as follows. We discuss  related work and Arabic linguistic facts in \S\ref{sec:related-work} and \S\ref{sec:arabic-backrground}, respectively. We describe the design and implementation of the web-based Arabic gender rewriter in \S\ref{sec:design} and conclude in \S\ref{sec:conclusion}.


\section{Related Work}
\label{sec:related-work}
Research has shown that NLP systems 
embed and amplify gender bias in a variety of core tasks such as machine translation (MT) \cite{rabinovich-etal-2017-personalized,elaraby2018,vanmassenhove-etal-2018-getting,escude-font-costa-jussa-2019-equalizing,stanovsky-etal-2019-evaluating,costa-jussa-de-jorge-2020-fine,gonen-webster-2020-automatically,saunders-byrne-2020-reducing,saunders-etal-2020-neural,stafanovics2020mitigating,savoldi2021gender,ciora-etal-2021-examining,savoldi-etal-2022-morphosyntactic, savoldi-etal-2022-dynamics} and dialogue systems~\cite{cercas-curry-etal-2020-conversational,dinan-etal-2020-queens,liu-etal-2020-gender,liu-etal-2020-mitigating,sheng-etal-2021-nice}. Most existing solutions to mitigate gender bias in NLP systems either focus on debiasing pretrained representations used in downstream tasks \cite{bolukbasi2016man,zhao-etal-2018-learning,manzini-etal-2019-black,zhao2020gender} or on training systems on gender-balanced corpora \cite{lu2018gender,rudinger-etal-2018-gender,zhao-etal-2018-gender,hall-maudslay-etal-2019-name,zmigrod-etal-2019-counterfactual}. 
More recently, text rewriting models were introduced to mitigate gender bias by either neutralizing the outputs of NLP systems or changing their grammatical genders to match provided users' gender preferences. \newcite{vanmassenhove-etal-2021-neutral} and \newcite{sun2021they} presented rule-based and neural rewriting models to generate gender-neutral sentences in English. For morphologically rich languages and specifically Arabic, \newcite{habash-etal-2019-automatic} and \newcite{alhafni-etal-2020-gender}, introduced gender identification and rewriting models to rewrite first-person-singular Arabic sentences based on the target user gender requirements. The task of gender rewriting was formally introduced by \newcite{alhafni-etal-2022-user} where they developed a new approach for Arabic gender rewriting in contexts involving two users (I and/or You) – first and second grammatical persons with independent grammatical gender preferences, and showed improvements over both \newcite{habash-etal-2019-automatic} and \newcite{alhafni-etal-2020-gender} systems. The tool we introduce in this work uses the best gender rewriting model developed by \newcite{alhafni-etal-2022-user}.\footref{alhafni22}


It is worth noting that our tool is similar to the recently introduced \texttt{Fairslator} \cite{mechura-2022-taxonomy}, a human-in-the-loop web-based tool for detecting and correcting gender bias in the output of MT systems translating from English to French, German, Czech, or Irish.\footnote{\url{https://www.fairslator.com/}} However, our work is different from theirs in the following ways:
\begin{itemize}
    \item \textbf{Input}: our system takes either Arabic or English sentences as an input, whereas \texttt{Fairslator} only handles English sentences.
    \item \textbf{Models}: the Arabic gender rewriter relies internally on both rule-based and neural models as opposed to \texttt{Fairslator}'s rule-based gender reinflection system. 
    \item \textbf{Evaluation}: the underlying gender rewriting model we use has been evaluated on  Arabic gender rewriting and post-editing MT output, and it achieves state-of-the-art results, whereas \texttt{Fairslator} was not evaluated on any of the four languages it targets.
    \item \textbf{Visualization}: we focus on visualization by highlighting Arabic gender-marking words in both the input and the output to provide a better user-experience. 
\end{itemize}




\section{Arabic Linguistic Background}
\label{sec:arabic-backrground}
Arabic has a rich morphological system that inflects for
gender, number, person, case, state, aspect, mood and voice, in addition to numerous attachable clitics (prepositions, particles, pronouns) \cite{Habash:2010:introduction}. Arabic nouns, adjectives, and verbs inflect for gender: masculine (\textit{M}) and feminine (\textit{F}), and for number: singular (\textit{S}), dual (\textit{D}) and plural (\textit{P}). Grammatical gender and number are commonly expressed using inflectional suffixes that represent some number and gender combination.  Pronominal clitics also express gender and number combinations, e.g.,  \<طبيبتكم> \textit{Tbyb+km} `your [masculine plural] doctor [feminine singular].
Gender and number participate in the morpho-syntactic agreement within specific constructions such as nouns and their adjectives and verbs and their subjects.

In practice, gender-specific words that are candidates for gender rewriting account for 10\% of all words in all sentences and 17\% of all words in gender-specific sentences.  These statistics are calculated from the Arabic Parallel Gender Corpus ({\APGC}) v2.1 \cite{alhafni-etal-2022-arabic}, which we use to train our models.





\section{Design and Implementation}
\label{sec:design}


\begin{table*}[ht!]
    \centering
        \begin{tabular}{p{0.48\textwidth} | p{0.48\textwidth}}
            \multicolumn{1}{c}{(a)} & \multicolumn{1}{c}{(b)} \\
            \vspace{-2pt} \includegraphics[width=0.48\textwidth]{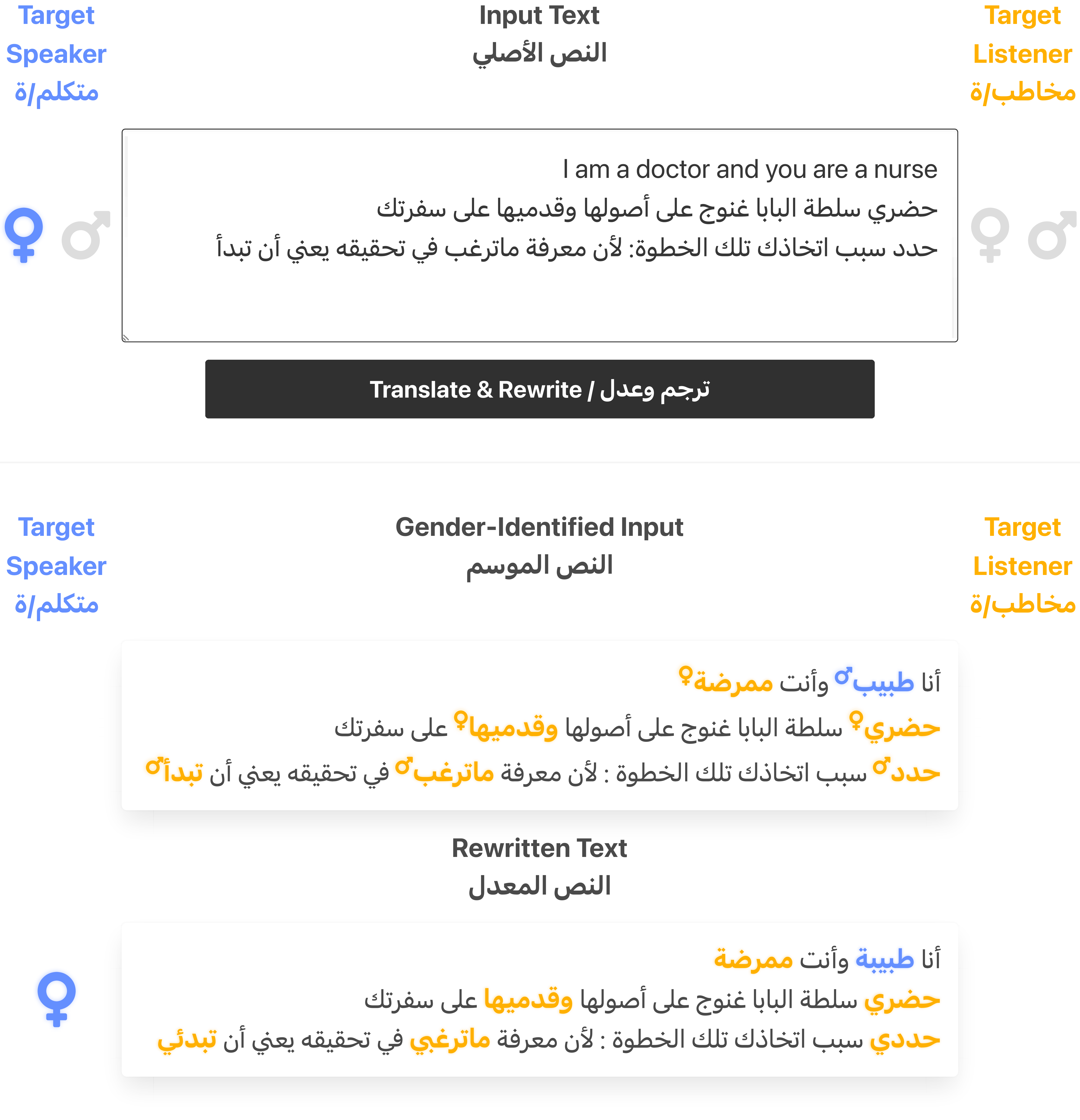} &
            \vspace{-2pt} \includegraphics[width=0.48\textwidth]{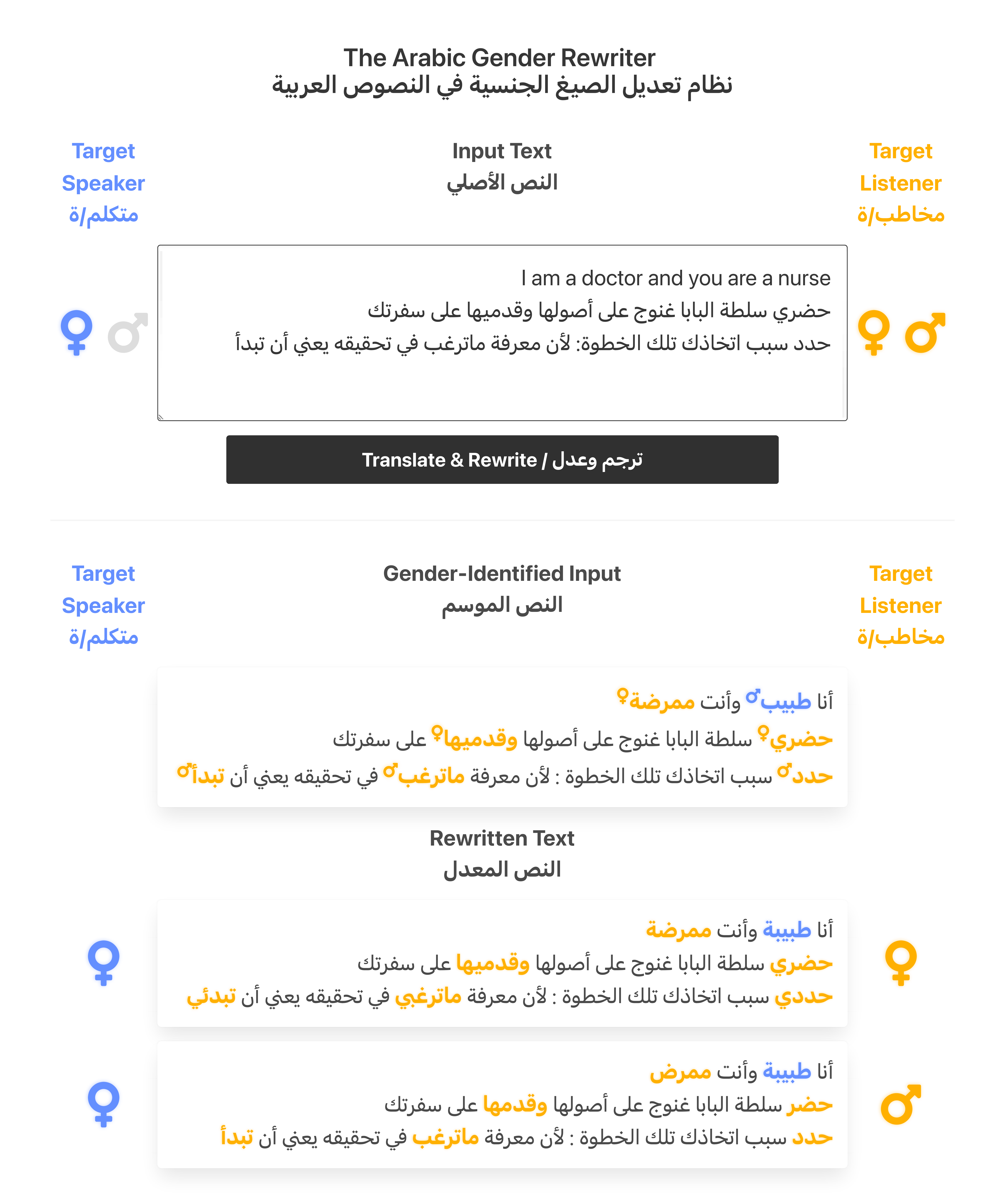}  \\\hline
            \multicolumn{1}{c|}{(c)} & \multicolumn{1}{c}{(d)} \\
            \vspace{-2pt} \includegraphics[width=0.48\textwidth]{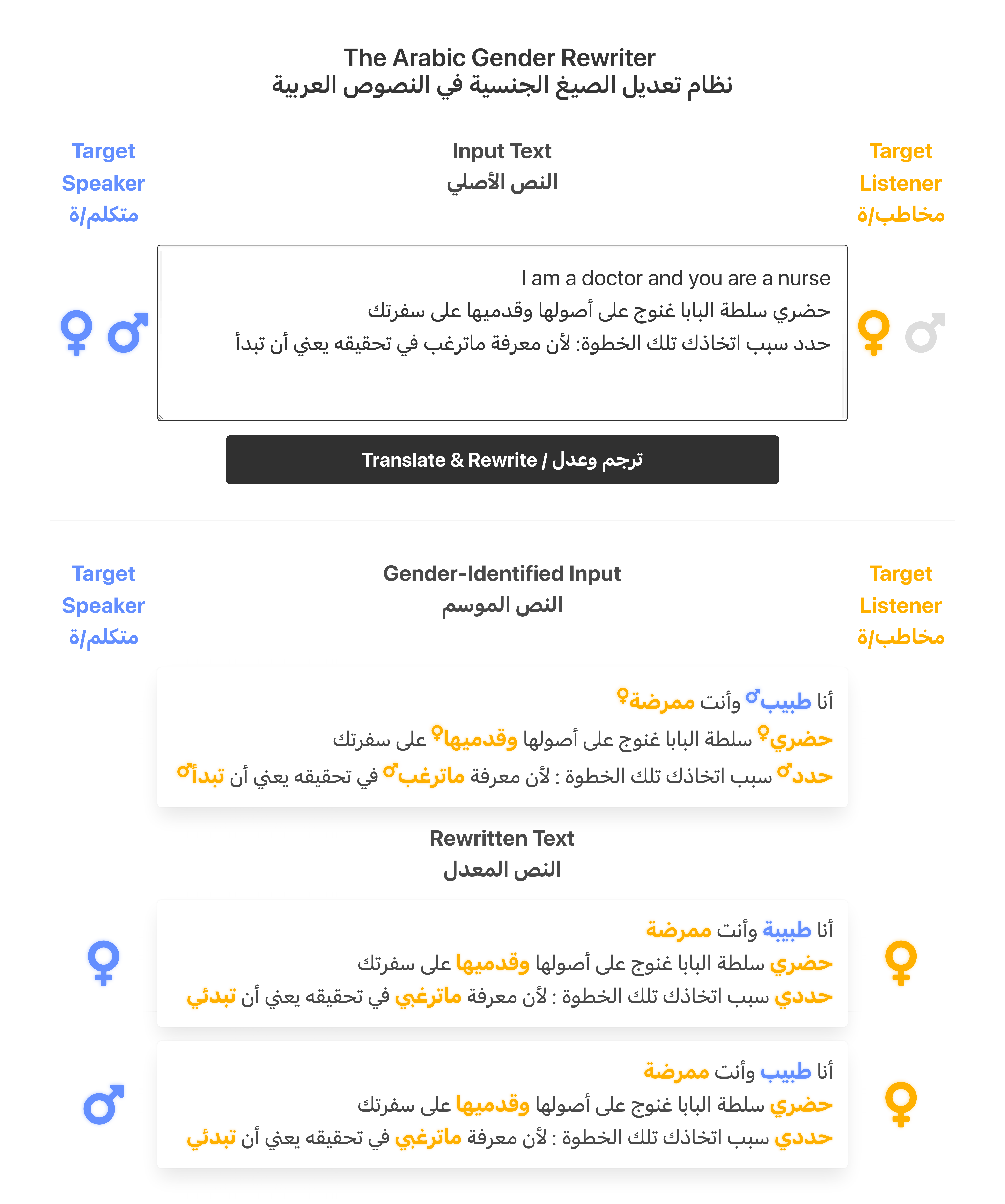} &
            \vspace{-2pt} \includegraphics[width=0.48\textwidth]{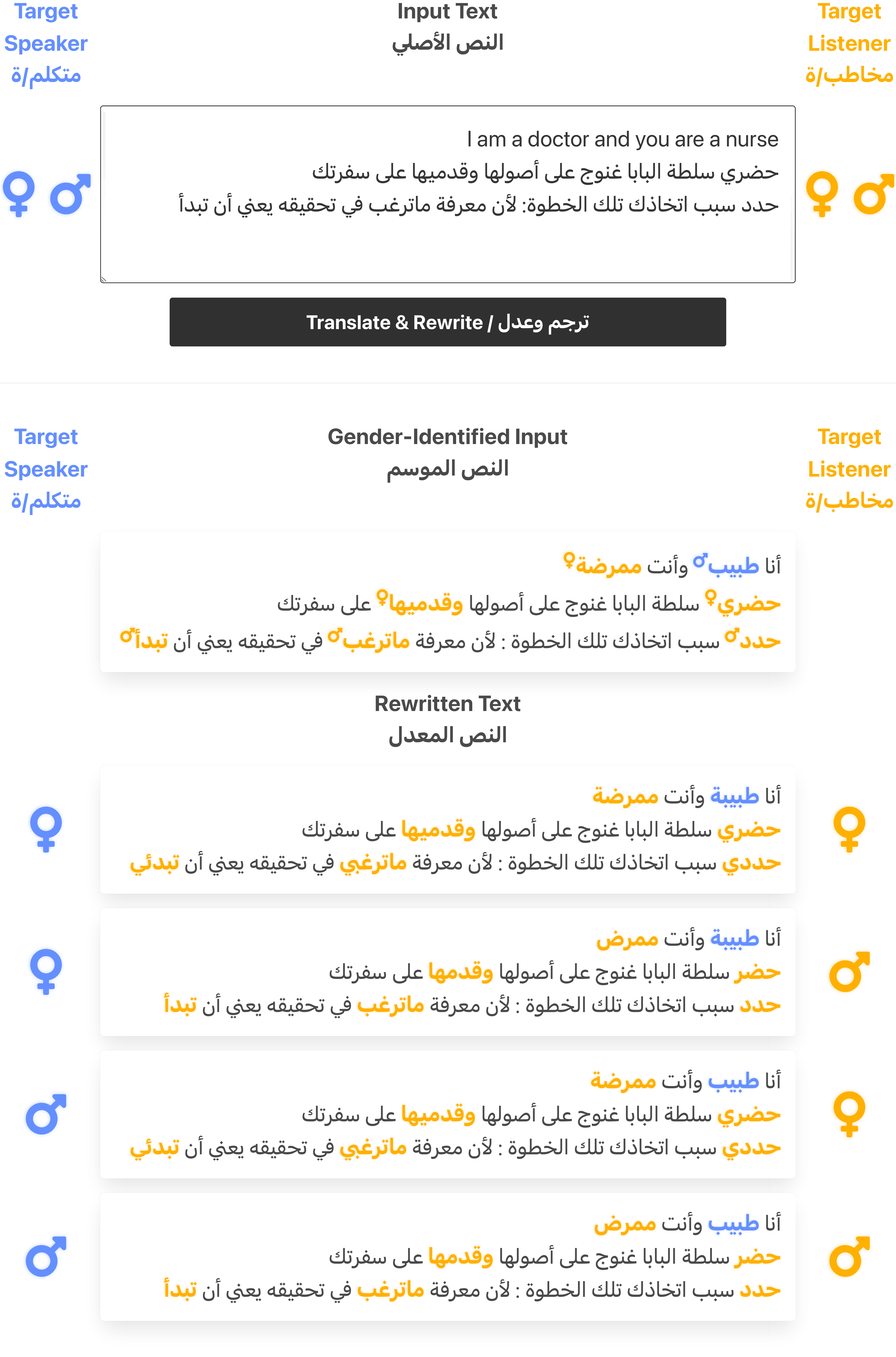} \\\hline
                
            \end{tabular}
    \captionof{figure}{The Arabic Gender Rewriter interface showing gender rewritten alternatives of three input sentences in four modes: (a) Target speaker $\female$ gender rewrites, (b) Target speaker $\female$ and target listener $\female$ and $\male$ gender rewrites, (c) Target speaker $\female$ and $\male$ and target listener $\female$ gender rewrites, and (d) Target speaker $\female$ and $\male$ and target listener $\female$ and $\male$ gender rewrites. Speaker gendered words are in \textcolor{blue}{\textbf{blue}} and listener gendered words are in \textcolor{orange}{\textbf{orange}}.}
    \label{fig:interface}
\end{table*}




\subsection{User Interface}
Our gender rewriting interface is publicly available at: \url{http://gen-rewrite.camel-lab.com/}. Figure~\ref{fig:interface}(a) shows the basic structure of the interface. At the top, there is a text box to input either English or Arabic text. At each side of the text box, there are two selection buttons to indicate the desired target gender preferences for the speaker and the listener ($\male$ is for masculine and $\female$ is for feminine). The user is able to select any possible combination of the desired target genders, including no target gender selection (i.e., requesting no rewriting). 

Once the user clicks on the \textit{Translate \& Rewrite} button, all input English sentences will be passed to Google Translate's API to translate them into Arabic before generating their gender alternatives. When the gender rewriting process is done, additional text boxes will appear: the first text box will always contain the gender-identified Arabic inputs and the rest of the text boxes will contain the gender rewritten alternatives. Each gender marking word in the gender-identified input text box will be labeled as either masculine ($\male$) or feminine ($\female$). First-person (i.e., speaker) gendered words are colored in \textcolor{blue}{\textbf{blue}} and second-person (i.e., listener) gendered words are colored in \textcolor{orange}{\textbf{orange}}.

The number of the text boxes containing the gender rewritten alternatives is based on the selected target gender preferences. Each one of those boxes will have a label at its sides indicating a particular target gender combination based on the users' selections. For instance, Figure~\ref{fig:interface}(a)  has one text box containing first-person feminine gendered alternatives of the input sentences. We discuss the screenshots in Figure~\ref{fig:interface} in more details in \S\ref{sec:examples}.














\paragraph{Front-end}
The front-end was implemented using \texttt{Preact}\footnote{\url{https://preactjs.com/}} for view control and \texttt{Bulma}\footnote{\url{https://bulma.io/}} for styling.

\paragraph{Back-end} The back-end was implemented in Python using \texttt{Flask} to create a web API wrapper for the gender rewriting model.\footnote{\url{http://flask.pocoo.org/}} We use the best performing gender rewriting model described in \newcite{alhafni-etal-2022-user}. The model was trained on the {\APGC} v2.1 in addition to augmented data from the OpenSubtitles 2018 dataset \cite{lison-tiedemann-2016-opensubtitles2016} and it consists of three components: gender identification, out-of-context word gender rewriting, and in-context ranking and selection. 

The gender identification component identifies the word-level gender label for each word in the input sentence. It leverages a word-level BERT-based \cite{devlin-etal-2019-bert} classifier that was built by fine-tuning CAMeLBERT MSA \cite{inoue-etal-2021-interplay}. Once the gender labels have been identified for each word in the input and given the desired users target genders, out-of-context word gender rewriting is triggered based on the compatibility between the provided users' target genders and the predicted word-level gender labels. The gender rewriting component employs three word-level gender alternative generation models in a backoff cascade setup: 1) Corpus-based Rewriter: a bigram maximum likelihood estimation lookup model; 2) Morphological Rewriter: a morphological analyzer and generator provided by CAMeL Tools~\cite{obeid-etal-2020-camel}; and 3) Neural Rewriter: a character-level sequence-to-sequence model with side constraints \cite{sennrich-etal-2016-controlling}. Since the three implemented word-level gender rewriting models are out of context and given Arabic’s morphological richness, this leads to producing multiple candidate gender alternative sentences. To select the best candidate output sentence, we rank all candidates in full sentential context based on their pseudo-log-likelihood scores \cite{salazar-etal-2020-masked}.

Results on the test set of {\APGC} v2.1 show that the best gender rewriting model achieves an M\textsuperscript{2}~\cite{dahlmeier-ng-2012-better} F\textsubscript{0.5} score of 88.42 and an average of 1.2 BLEU~\cite{Papineni:2002:bleu} increase when automatically post-editing Google Translate's output.


\subsection{Examples and Use Cases}
\label{sec:examples}

Figure~\ref{fig:interface} presents the different outputs of the gender rewriting tool for three input sentences, one in English and two in Arabic. The three sentences come from the examples presented in 
Figure \ref{fig:google-translate}, Figure~\ref{fig:baba-advice}(a), and Figure~\ref{fig:baba-advice}(b), respectively.

In Figure~\ref{fig:interface}(a), only the feminine target gender for the speaker is selected by the user. In this case, the system performs gender identification and then generates the first-person feminine gender alternative of the input sentences where all first-person masculine words are rewritten to feminine.  Figure~\ref{fig:interface}(b) shows an example where the feminine target gender for the speaker, and both the feminine and the masculine target genders for the listener are selected. In this case, the system outputs two gender rewritten alternatives for each input sentence, one for each selected target gender combination (i.e., speaker feminine -- listener feminine, speaker feminine -- listener masculine). Similarly, Figure~\ref{fig:interface}(c) shows an example where both the feminine and the masculine target genders for the speaker, and the feminine target gender for the listener are selected. Lastly, Figure~\ref{fig:interface}(d) is where all the target gender preferences are selected for both the speaker and the listener. In this case, the system generates all four possible gender rewritten alternatives for each input sentence.


\section{Conclusion and Future Work}
\label{sec:conclusion}
We introduced the User-Aware Arabic Gender Rewriter, a user-centric web-based system for Arabic gender rewriting in contexts involving two users. Our system takes either Arabic or English sentences as input, and provides users with the ability to specify their desired first and/or second persons target genders. The system outputs gender rewritten alternatives of the Arabic input sentences (or their Arabic translations in case of English input) to match the target users’ gender preferences. Moreover, the system highlights Arabic gender-marking words in both the input and the output to provide a better user-experience.

In future work, we plan to continue improving our gender rewriting back-end by adding better gender rewriting models and enhancing inference efficiency, as well as expanding gender identification and rewriting to third person entities. We also plan to improve the interface by enabling users to provide feedback that can be collected and used to enhance the performance of gender rewriting. We will also improve the visualization we use to highlight Arabic gender marking words by examining the added value it provides to different end users, from language learners to native text editors. \vspace{30pt}

\section*{Limitations and Ethical Considerations}
We acknowledge that by limiting the choice of gender expressions to the grammatical gender choices in Arabic, we exclude other alternatives such as non-binary gender or no-gender expressions. However, we are not aware of any
sociolinguistics published research that discusses such alternatives for Arabic.
We further recognize the limitations of the gender identification component we use in the back-end of the Arabic gender rewriter as it is based on a language model pretrained on a large monolingual Arabic corpus, which could possibly contain biased text. We realize the potential risks of maliciously misusing our system to intentionally produce gender alternatives that do not match the target users’ gender preferences; as well as the potential risks of negative reactions due to output errors. 

\bibliography{anthology,custom,extra,camel-bib-v2}
\bibliographystyle{acl_natbib}



\end{document}